\pdfoutput=1

\documentclass[11pt]{article}

\usepackage[final]{acl}

\usepackage{times}
\usepackage{latexsym}
\usepackage{float}
\usepackage[T1]{fontenc}

\usepackage[utf8]{inputenc}

\usepackage{microtype}
\usepackage{inconsolata}

\usepackage{graphicx}

\usepackage[utf8]{inputenc}

\usepackage{microtype}

\usepackage{{booktabs}}
\usepackage{multirow}
\usepackage{tcolorbox}

\usepackage{array}
\usepackage{caption}
\usepackage{subcaption}
\usepackage{todonotes}
\usepackage{xcolor}
\usepackage{booktabs}
\usepackage{algorithm}
\usepackage[noend]{algpseudocode}

\usepackage{graphicx}
\usepackage{tabularx}

\usepackage{xspace,mfirstuc,tabulary}

\newcommand{\instr}

\newcommand{\heart}{\ensuremath\heartsuit}

\newcommand{\club}{\ensuremath\clubsuit}

\usepackage{amsmath, amssymb, amsthm}

\usepackage{amsthm,amsmath,amsfonts,bm,xspace}
\usepackage{color}

\newcommand{\system}[1]{\textsc{#1}}
\newcommand{\data}[1]{\textsc{#1}}
\newcommand{\metric}[1]{\textsc{#1}}
\newcommand{\deleting}{deleting\xspace}
\newcommand{\obscuring}{obscuring\xspace}

\newcommand{\ourBenchmark}{\data{NaP$^2$}\xspace}
\newcommand{\echr}{\data{ECHR}\xspace}

\newcommand{\personaChat}{\data{PERSONA-CHAT}\xspace}

\newcommand{\privacyLeakage}{\metric{Privacy\_NLI}\xspace}

\newcommand{\rouge}{\metric{ROUGE-1}\xspace}
\newcommand{\rougeLsum}{\metric{ROUGE-Lsum}\xspace}
\newcommand{\bleu}{\metric{BLEU}\xspace}

\newcommand{\precision}{\metric{Precision}\xspace}
\newcommand{\acc}{\metric{Acc.}\xspace}
\newcommand{\fscore}{\metric{F1 Score}\xspace}

\newcommand{\oc}{\metric{OC}\xspace}

\theoremstyle{definition}

\newcommand{\gptf}{\system{GPT4}\xspace}
\newcommand{\gptfo}{\system{GPT-4o}\xspace}

\newcommand{\tfive}{\system{T5}\xspace}

\newcommand{\tfivesbase}{\system{T5-Base}\xspace}

\newcommand{\roberta}{\system{RoBERTa}\xspace}
\newcommand{\dpprompt}{\system{DP-Prompt}\xspace}
\newcommand{\dpmlm}{\system{DP-MLM}\xspace}
\newcommand{\disclosure}{\system{Disclosure-Abstraction}\xspace}

\newcommand{\flair}{\system{FLAIR-SCRUBBING}\xspace}

\newcommand{\llamatwo}{\system{Llama2-7B}\xspace}

\newcommand{\llamato}{\system{Llama3.1-8B}\xspace}

\newcommand{\legalbert}{\system{LEGAL-BERT}\xspace}

\newcommand{\ourmethod}{NaPaRe\xspace}
\definecolor{mgreen}{rgb}{0,0.7,0}

\newcommand{\cosine}{\system{Cosine Similarity}\xspace}
\newcommand{\finetune}{\system{Llama-alignment}\xspace}
\newcommand{\armo}{\system{ArmoRM}\xspace}

\def\eqref#1{(\ref{#1})}

\def\1{\bm{1}}

\def\vp{{\bm{p}}}

\def\vt{{\bm{t}}}
\def\vu{{\bm{u}}}

\def\vy{{\bm{y}}}

\DeclareMathAlphabet{\mathsfit}{\encodingdefault}{\sfdefault}{m}{sl}
\SetMathAlphabet{\mathsfit}{bold}{\encodingdefault}{\sfdefault}{bx}{n}

\newcommand{\privacyseg}{privacy segment\xspace}

\newcommand{\PrivacySeg}{Privacy Segment\xspace}
\definecolor{promptborder}{RGB}{169,169,169} 
\definecolor{prompttitle}{RGB}{105,105,105}  

\title{Zero-Shot Privacy-Aware Text Rewriting via Iterative Tree Search}
\author{Shuo Huang\textsuperscript{\rm \heart}, 
\textbf{Xingliang Yuan}\textsuperscript{\rm \club}, 
\textbf{Gholamreza Haffari}\textsuperscript{\rm \heart},
\textbf{Lizhen Qu}\textsuperscript{\rm \heart}\footnotemark[1], 
\\
\textsuperscript{\rm \heart} Monash University, \textsuperscript{\rm \club}University of Melbourne\\
\textsuperscript{\rm \heart}\{shuo.huang1, lizhen.qu, gholamreza.haffari\}@monash.edu,
\\\textsuperscript{\rm \club}xingliang.yuan@unimelb.edu.au}

\begin{document}
\maketitle
\footnotetext[1]{Corresponding author.}

\begin{abstract}

The increasing adoption of large language models (LLMs) in cloud-based services has raised significant privacy concerns, as user inputs may inadvertently expose sensitive information. Existing text anonymization and de-identification techniques, such as rule-based redaction and scrubbing, often struggle to balance privacy preservation with text naturalness and utility. In this work, we propose a zero-shot, tree-search-based iterative sentence rewriting algorithm that systematically obfuscates or deletes private information while preserving coherence, relevance, and naturalness. Our method incrementally rewrites privacy-sensitive segments through a structured search guided by a reward model, enabling dynamic exploration of the rewriting space. Experiments on privacy-sensitive datasets show that our approach significantly outperforms existing baselines, achieving a superior balance between privacy protection and utility preservation.


\end{abstract}

\section{Introduction}

The rapid integration of large language models (LLMs) into cloud-based applications has amplified privacy concerns, as user-generated texts often inadvertently disclose sensitive personal information. In domains ranging from healthcare~\cite{lison2021anonymisation} to legal proceedings~\cite{deuber2023assessing} and social media interactions~\cite{mireshghallah2023can}, the submission of unfiltered inputs to LLM APIs risks exposing details like medical histories, identities, or locations, which may be logged, analyzed, or misused by service providers. Traditional anonymization techniques, such as rule-based redaction or scrubbing, frequently compromise textual naturalness and utility, producing outputs that are awkward or semantically diminished. While finetuning LLMs on privacy-sanitized datasets~\cite{dou-etal-2024-reducing} mitigates some risks, this approach demands substantial computational resources and expertise, rendering it infeasible for individual users or resource-limited environments. Consequently, there is a pressing demand for zero-shot, open-source methods that enable local text rewriting, preserving privacy without sacrificing the coherence, relevance, and fluency essential for downstream applications.

Advancements in LLM-driven text anonymization have begun to address these challenges by leveraging generative capabilities to balance privacy preservation with utility and naturalness~\cite{huang2024nap, dou-etal-2024-reducing, staab2024large}. These techniques surpass the limitations of text-based differential privacy (DP) methods~\cite{du2023dpforward, meisenbacher2024dp}, which introduce noise to obscure identities but often result in degraded readability and task performance. Instead, they utilize accessible open-source models, such as T5~\cite{raffel2020t5} and LLaMA2-7B~\cite{dou-etal-2024-reducing}, to perform targeted paraphrasing that reworks sensitive elements while maintaining the original intent.

Despite these progresses, existing approaches remain constrained in three critical ways. First, they predominantly operate on predefined personally identifiable information (PII) categories or statically detected spans, offering limited adaptability to dynamic or user-specified privacy profiles---such as custom sensitivities to financial details in professional narratives or ideological nuances in public discourse\cite{huang2024nap}. Second, achieving robust results with open-source LLMs typically requires finetuning on specialized datasets, which are often unavailable or costly to curate for underrepresented domains. Third, by applying uniform strategies across all private elements, these methods overlook varying levels of information sensitivity---treating a casual mention of a hobby equivalently to protected health data---which leads to either excessive modifications that erode utility or inadequate protections that allow inference attacks, thus hindering precise calibration of the privacy-utility-naturalness trade-off.

We address these gaps with \ourmethod, a zero-shot, iterative tree-search algorithm for naturalness and privacy-aware text rewriting, deployable on medium-sized local LLMs to ensure fully offline operation. Formally, given a privacy specification $p$---encompassing PII lists, textual directives, or user profiles---and an input utterance $u$, NaPaRe generates an output $y$ that eradicates or conceals references to $p$, upholds non-sensitive content for effective cloud-based task execution, and mimics natural language with minimal semantic alterations. 

The pipeline of \ourmethod integrates precision and exploration in two phases. Initially, privacy segment alignment decomposes $u$ to pinpoint sensitive portions, calculating scores $Align_{t_j} = Pri(p, t_j)$ for each segment $t_j$ via embedding similarities or cosine metrics, isolating a sequence of targets ${t_p^{(1)}, \dots, t_p^{(m)}}$ for focused intervention. Subsequently, a Monte Carlo Tree Search (MCTS)-inspired framework~\cite{dainese2024generating} models rewriting as a decision tree: root nodes reflect partial states of $u$, with branches extending through actions---deletion for high-sensitivity spans or obscuration via generalization. Node selection employs Upper Confidence Bound for Trees (UCT) to weigh promising paths, while a controllable one-step LLM rewriter, prompted with privacy strategies, yields candidate sets $Y_{cand}$ gated by a thresholded utility function $LS(y, p_{seg}) \leq \gamma$. A reward model $R(y, p)$ synthesizes NLI-derived privacy entailment scores~\cite{huang2024nap} with domain-specific utility measures, enabling backpropagation to refine explorations iteratively. As outlined in Algorithm 1, the process advances sequentially: each segment's optimal rewrite, once threshold-compliant or budget-exhausted, proceeds to the next iteration, culminating in a cohesive $y_{final}$.

This structured, reward-guided search distinguishes \ourmethod by dynamically navigating the rewriting space, supporting flexible privacy handling without finetuning.  It ensures the searching to generate high quality rewrite even with less capable model compared to commercial LLM like \gptfo Our contributions include:
\begin{itemize}
\item \ourmethod: An innovative, tree-based iterative rewriter that fuses sampling, UCT selection, and composite rewards to explore deletion and obscuration strategies, adaptable to diverse $p$ in zero-shot settings.
\item Comprehensive evaluations on the \ourBenchmark corpus and \echr legal judgments, measuring privacy (via \privacyLeakage and PII F1), utility (ROUGE-1 and judgment accuracy), and naturalness (perplexity and human assessments). \ourmethod yields a 22.3\% relative privacy enhancement over baselines, with negligible utility drops and fluency within 1.5 perplexity points of originals, outperforming redaction tools and LLM paraphrasers.
\end{itemize}
\begin{itemize}
    \item We propose \ourmethod, a tree-based iterative privacy-aware rewriting algorithm inspired by Monte Carlo Tree Search (MCTS)~\cite{dainese2024generating}, explores rewriting strategies through a structured decision-making process that combines repeated sampling, reward-based filtering.
    \item We conduct extensive experiments across three dimensions: privacy leakage, utility (measured via task-specific semantic preservation metrics), and naturalness (assessed through perplexity and human evaluation). \ourmethod achieves a 22.3\% relative improvement in privacy protection with minimal utility loss, while maintaining fluency within 1.5 perplexity points of the original sentence on average, outperforming both redaction-based approaches and competitive LLM-based rewriting methods.
\end{itemize}
\begin{figure*}
    \centering
    \includegraphics[width=1\linewidth]{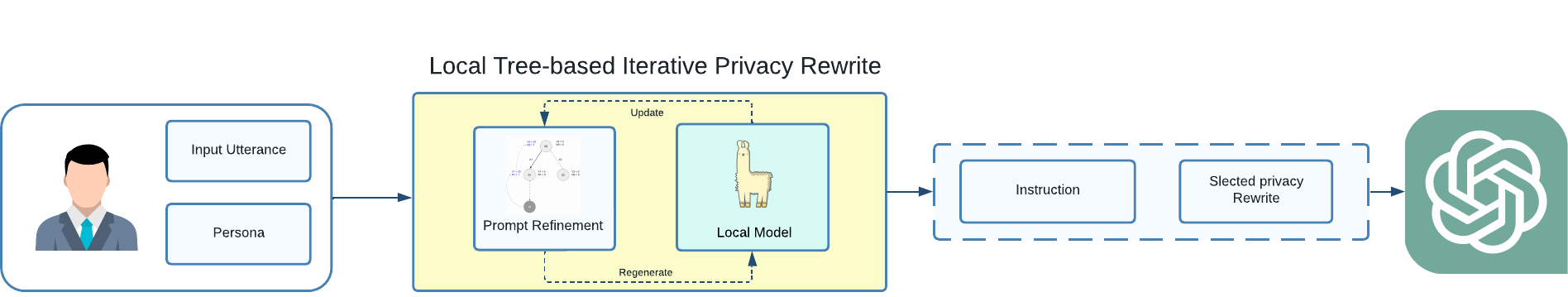}
    \caption{Tree Search-enhanced Iterative Privacy Rewrite works as intermediate layer to rewrite the textual input from user to remove private information provided by persona.}
    \label{rewrite_workflow}
\end{figure*}

\begin{figure*}
    \centering
    \includegraphics[width=1\linewidth]{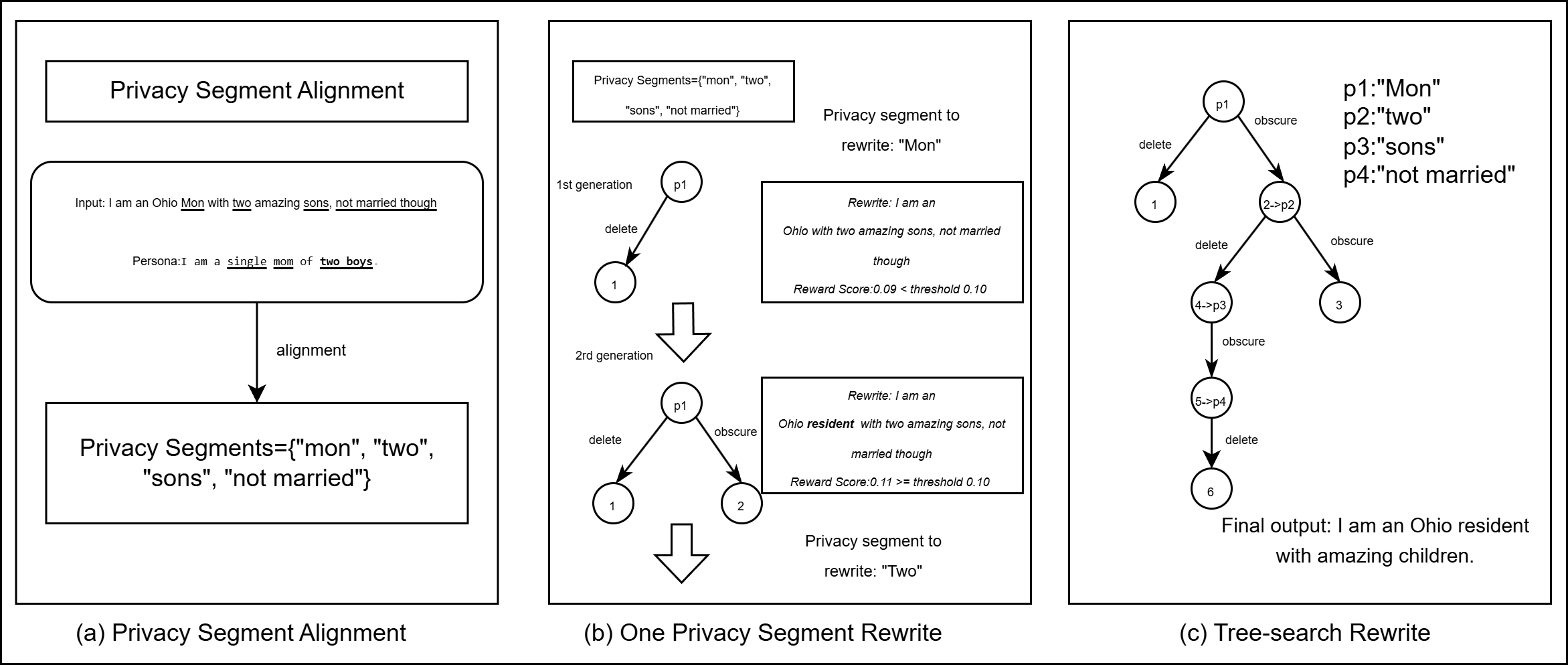}
    \caption{Rewrite example and full rewrite pipeline}
    \label{rewrit_example}
\end{figure*}

\section{Privacy-Aware Text Rewriting}
\paragraph{Task formulation}

Given the privacy information described $\vp$ from the user, the objective of our method is to leverage the generation ability of locally deployed LLM to rewrite the input utterance $\vu$ in order to either remove or obscure any private information presented in $\vp$. The $\vp$ is generalized privacy specification of a user. It can be a set of PII removed, text-formatted privacy requirements or profiles. The generated sentence $\vy$ should satisfy the following requirements:
(1) $\vy$ does not reveal any private information identified in $\vp$. (2) The rewritten sentence $\vy$ maintains the non-private content in $\vu$ such that the resulting rewritten sentence can perform the proper tasks in the cloud.
(3) The generated sentence does not warn the untrusted party that the text has already been rewritten(preserving the naturalness of the generated sentence).

\paragraph{Assumptions} 
To avoid uploading the private information to the cloud services, our rewriting model is locally deployed, and the inference is completely offline. We assume that for the single user, our method works as an application on the local device. The private information can be either a predefined set of attributes like location, gender identity or marriage status or any arbitrary information the user typed in as private information $\vp$ on their own devices\cite{dou-etal-2024-reducing}. 

\subsection{One Step LLM Text Rewrite}
We propose a controllable rewriting mechanism, $Rewrite$, that generates privacy-preserving text rewrites. The rewrite proceeds as follows: Given an input sentence $x$ and one privacy segment $p_{seg}$, $Rewrite$ first uses a stochastic language model $G_{LLM}$ with a privacy-aware prompting strategy $a\in \{obscuring,deleting\}$ to produce $N$ candidate rewrites $\mathcal{Y}_{cand}$. Each candidate $y\in \mathcal{Y}_{cand}$ is scored by a utility function $L_S(y, p_{seg}) \in [0,1]$, quantifying the residual presence of private attributes or quality of generation varied on the monitor function applied. A monotonic threshold is set $\gamma$, which defines how a generation will be accepted. Higher the threshold, the  Candidates with $L_S(y, p_{seg}) \le \gamma$ are retained in an acceptable set $\mathcal{Y}{acc}(x)$. One example will be randomly selected from this accepted set. If no candidates meet this criterion, the mechanism returns the sentence with the highest utility score from $\mathcal{Y}_{cand}$. In the middle of tree generation, we consider the same threshold value for the $\gamma$ mentioned in the one-step text rewrite.




\begin{algorithm}
\caption{Tree-Structured Iterative Privacy Refinement}
\label{alg:tree-search}

\begin{algorithmic}[1]
\Statex \textbf{Input:} Input sentence $x$. Reward model $Reward$, Rewrite strategy set $A=\{deleting,obscuring\}$. One Step Privacy Rewrite Algorithm $Rewrite$. Tree Generation Budget $B$, Sampling Budget $C$
\Statex \textbf{Output:} Privatized Sentence.

\State Extract \privacyseg $T_p = \{\,t_{p}^{(1)}, t_{p}^{(2)}, \ldots, t_{p}^{(m)}\}$ from $x$ according to $p$
\State Initialize root state $s_0 \gets x$

\For{each \privacyseg $t_{p}^{(i)}$ in $T_p$}
    \State Initialize a new search tree with root node $s_0$, $t_{p}^{(i)}$ 
    \For{$k = 1$ to $B$} 
        \State \textbf{Selection:} Traverse the tree from the root, selecting child nodes via UCT to select a leaf node with action $a\in A$ with UCT probability.
        \State \textbf{Evaluation:} For each newly created child node, Use the generated sentence of parent node to produce the updated sentence at that node.$y' =  Rewrite(x,a,C)$
        \State \textbf{Compute the reward} $r$ by passing the node’s sentence into the reward function $\mathcal{R}$. $r \gets \mathcal{R}(y',t_{p}^{(i)})$
        \State \textbf{Backpropagation:} Propagate the reward $r$ up the tree, updating $Q(\cdot)$ and visit counts $N(\cdot)$ for each ancestor node.
        \If {$r' \geq \gamma$ }
            \State Break
        \EndIf
    \EndFor

    \State traverse leaf node to best leaf node $leaf_{best}$ and $y_{t_{p}^{(i)}} \gets Rewrite(leaf_{best},\epsilon)$
    
    \State Then set $s_p \gets  y_{t(i)_p }$ as input sentence  for the next private token.
\EndFor

\State When generation finished we will set the last generated example as our final output $y_{final} \gets y_{t_{p}^{(m)}}$
 
\State \textbf{Return:} $y_{final}$
\end{algorithmic}
\end{algorithm}

\subsection{Tree Search Iterative Refinement Privacy Rewrite}
Protecting personal information in text requires precise and context-aware rewriting rather than generic text obfuscation. Existing approaches to privacy-aware text generation often focus on named entity masking \cite{lison2021anonymisation} or sentence-level paraphrasing \cite{dou-etal-2024-reducing}, but these methods can either lead to excessive content removal or fail to fully obscure sensitive details. To address these challenges, we propose a tree-search-based rewriting framework that instructs the model to explicitly rewrite \privacyseg within an utterance in zero-shot manner.
Our approach follows a structured two-stage pipeline:

\textbf{\PrivacySeg Alignment}: A span alignment process maps privacy-sensitive spans in the input utterance to the semantic attributes of a persona description. This step ensures that rewriting actions are applied to the most relevant parts of the text, improving precision and control. \\
\textbf{Tree-Search Rewriting}: A stepwise decision-making process selects different rewrite strategies to perform privacy-preserving rewrite. A reward function evaluates the effectiveness of each rewrite, allowing the model to refine outputs iteratively.

\subsubsection{\PrivacySeg Alignment}
Directly prompting an LLM to remove personal information is often unreliable, as models struggle to identify and modify implicit disclosures \cite{staab2023beyond} especially in the scenario that one sentence contains multiple private information to rewrite. Instead, we consider this as a privacy segment alignment strategy with given private specification to decompose given input sentence and perform rewrite step by step.
We define a mapping function that selects \privacyseg from the input utterance $\vu$ to align information in the persona $\vp$. From the semantic of $\vp$, we identify the corresponding segment $t_s$ in $\vu$ that has the highest alignment score. Such score can be measured using similarity metrics such as cosine similarity or finetuned language model.

Formally, for each segments $\vt_j \in \vu$, we compute:

\begin{equation} Align_{\vt_j} =  \text{Pri}(\vp, \vt_j), \end{equation}

where $\text{Pri}(\vp,\vt_j)$ denotes the private alignment score between tokens $\vt_j$ and persona $\vp$. This mapping creates a set of aligned token ${ (t_1,t_2 ...t_m) }$, which identifies the specific tokens in $\vu$ that are likely to reveal private information.

\subsubsection{Tree-Search Privacy Rewriting}
To rewrite each \privacyseg with different strategies iteratively, we model the rewriting process as a tree-search problem, where each node represents a modified version of the sentence, and branches correspond to different rewrite actions applied to a single \privacyseg.
\paragraph{Action Space.}
At each node (i.e., each intermediate rewrite state), the algorithm considers two possible \emph{rewrite strategies} for a single \privacyseg. Concretely:
\begin{itemize} 
\item\textbf{\deleting}: Remove the \privacyseg from the sentence. 
\item \textbf{\obscuring}: Replace the \privacyseg with a less specific or more general term. 
\end{itemize}
\paragraph{UCT}
From a given node, we use Upper Confidence bounds applied to Trees (UCT) \cite{kocsis2006bandit} to calculate reward and adjust probability for each path. The equation and explanation of UCT can be found in Appendix.~\ref{app:uct}.

\paragraph{Reward.}
A Reward function $\mathcal{R}$ monitors each candidate rewrite and outputs a $r$ that reflects the level of privacy or quality of rewrites achieved. Formally, for a rewrite $y$, the reward function returns
\[
    r \;=\; \mathcal{R}\bigl(y,p\bigr),
\]
which we compare against a threshold $\gamma$. If $R(y) \ge \gamma$, we consider the rewrite acceptable (the node is ``good enough''); otherwise, further rewriting is needed. 

\paragraph{Algorithm.}
Algorithm~\ref{alg:tree-search} outlines our procedure. We initialize the tree with a \emph{root node} corresponding to the original sentence $x$. We select a single \privacyseg to rewrite and \emph{uniformly} sample one of the two rewrite strategies for that segment at the first step. After one generation step, the discriminator evaluates its reward $r$. The procedure then:

\begin{enumerate}
    \item \textbf{Update Node Reward and re-weight:} If $r \leq \gamma$, the generation continues and propagates its score back up the tree. Precisely, we return to the root node or a higher-level branch. We \emph{re-weight} the probability of choosing each rewrite strategy based on observed rewards. For new node root expansion, we sample from the root node to the new leaf based on the updated probability.
    \item \textbf{Termination Check:} If any leaf node exceeds the reward threshold $\gamma$, the algorithm terminate the generation for current segment. Alternatively, if \emph{computation budget} is reached without finding a suitable rewrite, we will traverse the leaf node to get the best generation so far.
\end{enumerate}
For each rewrite, we adopt our one-step text rewrite with some modifications. We will consider reward model in tree search as our monitor function.
Once the best rewrite is identified for the first \privacyseg, we fix that segment's transformation and proceed to the next \privacyseg, treating the partially rewritten sentence as the new root. This process continues until all \privacyseg in $x$ are processed.

\section{Experiments}

\subsection{Experimental Setup}
\paragraph{Datasets}

\underline{\textit{\ourBenchmark}} The Naturalness and Privacy-Preserving Rewriting Corpus (\ourBenchmark), based on the open-ended dialogue corpus \personaChat~\cite{zhang2018personaChat}, is designed to enable machines to adopt privacy-preserving rewriting strategies similar to those used by humans, specifically focusing on strategies like deletion and obscuration. The dataset provides persona as a privacy specification for rewriting, with curated human rewrites as targets.

\underline{\textit{\echr}}~\cite{chalkidis2019neural} \echr is an English legal judgment prediction dataset containing cases from the European Court of Human Rights (\echr) with full descriptions of defendants’ personal information. We followed the PII definition and tagging method from Flair NER, as done by ~\citet{lukas2023analyzing}. We consider PII as the privacy specification for rewriting. We sampled a test set with 298 examples to evaluate the utility of the rewrite and the baseline methods in a legal judgment prediction task. Each record contains raw text, a masked sentence (processed by the Flair NER tagger), a corresponding list of masked words, and the entity class removed.

\paragraph{Baselines.} 
For our rewriting model, we use \llamato to demonstrate the effectiveness of our approach. For the reward function in our tree-search algorithm, we employ \armo~\cite{wang2024armo}, a state-of-the-art (SOTA) reward model that evaluates generation quality based on the prompt. We use the generated score from this model as the reward score for each generation step. Detailed introduction of \armo can be found in Appendix.~\ref{app:armo}. We also considered the \privacyLeakage as our reward function and a combined approach of both models. The comparison is discussed in 
section 4. 
To comprehensively evaluate our method,we compare it with several highly relevant baseline models:
\dpmlm~\cite{meisenbacher2024dp} and \dpprompt~\cite{utpala2023locally}: Differentially private text rewriting methods that utilize masked language models (MLM) and zero-shot, prompt-based rewriting, respectively, to enhance utility. \disclosure: A SOTA privacy abstraction model fine-tuned on \llamatwo. \flair scrubbing: Used as a baseline for \ourBenchmark and as a privacy-segmentation alignment model for \echr, as it is commonly used in commercial PII detection and extraction tools. GPT-4\cite{achiam2023gpt} is used for zero-shot rewriting to compare our approach with state-of-the-art general-purpose models. \tfivesbase-\ourBenchmark: Adapted from ~\citet{huang2024nap}, which achieves the best performance on this dataset so far.

The implementation detail and hyperparameter setting can be found in Appendix.~\ref{appendix::baseline}

\paragraph{Evaluation Metrics.}

\underline{\textit{Privacy.}} For accessing the privacy leakage of the rewritten sentence, we adopted the automatic privacy evaluation \privacyLeakage~\cite{huang2024nap}. It utilizes the \roberta model trained on the Multi-Genre Natural Language Inference (MNLI) corpus~\cite{mnli} to assess the extent to which personal information in personas can be inferred.A higher metric indicates greater preservation of private information. For \echr, we also consider the PII scrubbing success rate for rewritten sentences, we report the matching score using \precision and \fscore of extracted PII after rewrite.

\underline{\textit{Utility.}} For utility evaluation, we use the respective metric for \ourBenchmark and \echr. For open domain generation of \ourBenchmark, We adopted the metric \rouge from \cite{dou-etal-2024-reducing} to encourage the generation diversity meanwhile, we use rank the examples trained from baseline to assess the diversity. And for \echr, we consider the downstream task of legal judgment prediction with accuracy(\acc) and \fscore.

\underline{\textit{Naturalness.}} 
For naturalness of sentence, following previous work to measure the smoothness and naturalness of generated sentence~\cite{pan2024unsupervised}, we used Perplexity(PPL) computed by GPT2.

\section{Security Analysis}

Despite advances in privacy-aware text rewriting, a critical vulnerability persists: adversaries aware of the rewriting algorithm may attempt to reconstruct original sensitive information from sanitized outputs. Motivated by this risk, which could undermine \ourmethod's zero-shot, local deployment for protecting user data in cloud interactions, we evaluate robustness against theoretically optimal reconstruction attacks inspired by text sanitization vulnerabilities~\cite{tong2025vulnerability}. Using medium-sized LLMs like \llamatwo for rewriting, we adopt frameworks for context-free and contextual Bayesian attacks, deriving Attack Success Rate (ASR) bounds at the token level.

For the context-free optimal reconstruction, the adversary recovers tokens via:
\[
x_i' = \arg \max_{x_i' \in X} \frac{\Pr(y_i | x_i') \Pr(x_i')}{\Pr(y_i)},
\]
Where $X$ is the sensitive token set, $y_i$ is the rewritten token, and probabilities reflect prior distributions and rewriting mechanisms. Contextual variants incorporate adjacent tokens $c_i$. We focus on differing tokens between input $u$ and output $y$, applying token-to-token alignment for length variations.

\subsection{Results and Discussion}

\paragraph{\ourBenchmark}
The detailed evaluation results are shown in Table.~\ref{tab:baseline}

\underline{\textit{Privacy.}}
Privacy is quantified using \privacyLeakage, which determines weather rewritten sentence can entail the privacy information provided. We set the \ourBenchmark-Human Rewrite as a baseline for a more clear comparison of each method. \ourmethod achieves a Privacy-NLI score of 93.02\%, achieving competitive privacy preserving ability with a tuned model. This indicates that our approach effectively modifies \privacyseg while ensuring the rewritten text aligns with privacy constraints. 

\underline{\textit{Utility.}}
To measure utility, we report \ourBenchmark-Human Rewrite as a reference. Unlike row-wise comparisons in the table (which evaluate generated output against human rewrites), this metric calculates the overlap between the original input and human rewrite targets. Our method achieves results closest to the original input while maintaining high privacy preservation scores. Importantly, our method does not require additional model fine-tuning. Under high privacy-preserving constraints, \rouge and \bleu compute the overlap between generated results and human rewrite references. Our method achieves 73.68\% in \rouge. We aslo provided extra utility metrics for open-ended generation tasks detailed in Appendix~\ref{app:utility-metrics}

\underline{\textit{Naturalness.}}
Naturalness is assessed using Perplexity (PPL), where lower values indicate more fluent and human-like text. Our model achieves a PPL of 151.83, comparable to human rewrites. GPT-4 achieves the lowest PPL, indicating more favorable generation fluency in the test model. Extremely high PPL suggests that rewritten text may contain unnatural or irrelevant modifications, as seen in \dpmlm and \dpprompt, both of which exceed 788 PPL. We also test the naturalness scored by GPT-4o. It shows consistency with PPL tested. Noted that scores from \llamatwo and \gptf shows best in PPL and LLM score in naturalness. And \ourmethod shows the closest naturalness with human rewrites. 
\paragraph{\echr} 
For \echr we report a separate table for privacy and naturalness in Table. \ref{tab::echr_privacy} and for utility we report the predicted results for finetuned \legalbert model\cite{chalkidis-etal-2020-legal} in Table.~\ref{tab:echr_prediction}. 

\underline{\textit{Privacy.}}
Unlike \ourBenchmark, the \privacyseg for \echr is extracted using \flair, following the approach of \cite{lukas2023analyzing}. After performing privacy rewriting, we test whether PII can still be extracted using \flair. We report the matching scores between the predicted and ground truth PII sets using \precision, \fscore, and \rouge. Since there is no human-labeled ground truth, we only compare privacy performance against \dpprompt and \dpmlm. A higher score indicates higher overlap between predicted and ground truth PII, meaning more privacy leakage. After rewriting, \ourmethod achieves only 4.79\%, the lowest among all methods, proving better privacy preservation. 

\underline{\textit{Utility.}}
For utility, we adopt legal judgment prediction \cite{chalkidis2019neural} as a downstream task for \echr. Given a legal case, the model predicts a binary judgment outcome. We use a \legalbert model \cite{chalkidis-etal-2020-legal} fine-tuned on the \echr training set to measure utility changes. As a reference, we include results from original test inputs and \flair outputs. Since \legalbert has a strict token limit, we evaluate two settings: full rewrite (average 17 sentences per case) and partial rewrite (only 5 sentences per case).As shown in Table~\ref{tab:echr_prediction}, \dpprompt performs competitively with our method under the 5-sentence rewrite setting. However, in the full rewrite setting, performance drops significantly, suggesting that excessive rewriting introduces greater diversity, impacting final predictions.

\underline{\textit{Naturalness.}}
Higher PPL scores confirm our observations—sentence-by-sentence rewriting introduces inconsistencies, especially across long legal cases, resulting in higher PPL values.

\begin{table*}
\centering
\resizebox{0.9\textwidth}{!}{
\begin{tabular}{l |rrrrcrr}
\toprule
 Method &  \privacyLeakage  &  \rouge &   \rougeLsum  &\bleu & PPL  &LLM\\

\midrule
\ourBenchmark-Human Rewrite &         \textbf{92.59\%}          &    90.90\%              &       90.90\%               &   77.10        &\textbf{118.1}   &     3.86      \\
\dpmlm    &             79.16\%&         45.05\% &                           45.46\%  &0.3956     &        1108.28    &  1.39    \\
\dpprompt    &             77.65\% &        85.71\% &                          57.14\%  &           14.79   &   788.99 & 1.00  \\
\flair    &              86.14\% &          53.33\% &                           53.33\%  &         48.68      & 202.46  & 2.97 \\
\disclosure &     65.30\%       &    21.05\%       &      21.05\%             &   4.01         &          77.21 &    	\textbf{4.77}     \\

\gptf &       82.24\%           &    33.33\%        &     33.33\%   &  9.88  & 83.35  &      4.18       \\

\tfive-\ourBenchmark-\gptf & 93.81\%          &  73.01\% & 72.78\%   & 37.47  & 279.35  &      	3.00       \\

\ourmethod-\llamato &                  \textbf{93.02\%}         &     \textbf{ 73.68\%} &    73.68\%     &             25.03 &   \textbf{151.83} &      4.0        \\

\bottomrule
\end{tabular}}
\caption{overall Evaluation on \ourBenchmark. We use \privacyLeakage to evaluate the privacy preservation of target private specification. LLM indicates the average naturalness score by GPT-4o. Detailed template and explaination can be found in Appendix.~\ref{app:naturalness}   }\label{tab:baseline}
\end{table*} 


\subsection{Ablation Study for Tree Search Method}

\begin{table}[t]
\centering
    \resizebox{\columnwidth}{!}{  
        \begin{tabular}{l| p{1.8cm} p{1.8cm} p{1.8cm}}  
            \toprule
 Method &  \fscore &  \rouge &   PPL \\

\midrule
\dpprompt    &           15.68\%    &     16.19  \%     &          279.35                               \\
\dpmlm    &          16.46\%     &      15.78\%      &                            590.26               \\

\ourmethod-\llamato &          5.18\%                        &        4.79\%               &     680.45    \\

\bottomrule
\end{tabular}}
\caption{Privacy and naturalness measurement for \echr}\label{tab::echr_privacy}
\end{table}


\begin{table}[t]  
    \centering
    \resizebox{\columnwidth}{!}{  
        \begin{tabular}{l | p{1.5cm} p{1.5cm} p{1.5cm}}  
            \toprule
            Method & \acc & \precision & \fscore  \\
            \midrule
            Original Input &  82.00\%   &      90.00\%     &    85.71\%        \\
            \dpmlm &34,60\% & 100.0\% &14.47 \% \\
            \dpprompt & 58.19\% & 95.69\% & 58.74\%\\
            \flair & 29.76\% & 0.00\% &  0.00\%  \\
            \ourmethod-5 &  68.00\% &94.73\%  &69.23\% \\
            \ourmethod-all &   48.00\% & 100\% &     35.00\%         \\

            \bottomrule
        \end{tabular}
    }
    \caption{Evaluation on Legal Judgment Prediction}
    \label{tab:echr_prediction}
\end{table}

To further validate our method and provide empirical justification, we conducted an extensive ablation study on \ourBenchmark to evaluate different settings and design choices for our approach.

\begin{table}[t]
\centering
    \resizebox{\columnwidth}{!}{  
        \begin{tabular}{l  | p{1.8cm} p{1.8cm} p{1.8cm}}  
            \toprule
 Method &  \privacyLeakage &  \rouge  &   PPL \\

\midrule
One-step &      61.02 \%              &         45.16 \%        &        48.09        \\
Random    &              92.23\% &         52.17\% &                            5817.24  \\
Greedy   &              95.09\% &         35.29\% &                            405.99  \\
Chain &                  91.43\%             &       34.48 \%      &         251.61        \\
\ourmethod   &              93.02\% &          73.68\% &               151.83         \\

\bottomrule
\end{tabular}}
\caption{Multi-step verification of tree search generation}\label{tab::step_verification}
\end{table}

\begin{table}[t]
\centering
    \resizebox{\columnwidth}{!}{  
        \begin{tabular}{l | p{2cm} p{2cm} p{2cm}}  
            \toprule
 Method &  \fscore &  \rouge &   \oc \\

\midrule
\cosine    &              43.05\% &          72.02\% &                            43.05\%          \\
\armo   &             37.50\% &         68.35\% &                            40.33\%         \\
\finetune    &             37.74\% &        89.38\% &                            88.88\%        \\

\bottomrule
\end{tabular}}
\caption{Evaluation for alignment options}\label{tab::alignement_option}
\end{table}


\begin{table}[t]
\centering
    \resizebox{\columnwidth}{!}{  
        \begin{tabular}{l p{1.8cm} p{1.8cm} p{1.8cm}}  
            \toprule
 Method &  \privacyLeakage &  \rouge &   PPL \\

\midrule
\armo &        93.02\% &          73.68\% &    151.83        \\
\privacyLeakage   &      94.88\% &         30.00\% &        132.32       \\
combined  &              88.55\% &         36.36\% &                            99.23             \\

\bottomrule
\end{tabular}}
\caption{Comparison of Tree search discriminator function}\label{tab::discriminator}
\end{table}

\paragraph{RQ 1: Is Multi-step tree-based improvement better than single-step rewriting?}
Our proposed method follows a multi-step rewriting approach, which is not necessarily superior to other settings. To investigate this, we conducted experiments in \llamato with different rewrite settings:
\begin{itemize}
    \item \textbf{One Step} This is the setting that we only perform our sentence-level privacy rewrite in one step with the provided privacy specification.
    \item \textbf{Random} For random, we consider not using UCT to update the reward and let the tree expand randomly with the same computation budget.
    \item \textbf{Greedy} We asked the model to expand in one route with multiple rounds until it reached the computation budget or satisfied the reward threshold.
    \item \textbf{Chain} In this setting, we consider one time rewrite for each private token aligned by privacy specification to form a rewrite chain rather than multi-step refinements. 
\end{itemize}
As shown in ~\ref{tab::step_verification}, the one-step rewrite performs the worst in privacy preservation but achieves the lowest PPL, as only a single rewrite is performed. Random rewriting has high \privacyLeakage along with the highest PPL, indicating the importance of controlled generation. Greedy achieves the highest \privacyLeakage but tends to overwrite sentences, resulting in a low \rouge score. Chain generation suffers from a similar issue, though it does not achieve as high \privacyLeakage as \ourmethod.

\paragraph{RQ 2: What is the best choice for discriminators?}
When designing the reward model as discriminator, we considered options that can monitor the generation quality of our model. To avoid increasing the computational burden, we did not employ fine-tuning-based reinforcement learning and instead focused on existing models and functions. We primarily considered three settings: using \privacyLeakage, reward model \armo, and a combination of both to evaluate the rewrite.
As shown in Table~\ref{tab::discriminator}, using \privacyLeakage for each \privacyseg achieves the highest overall \privacyLeakage on persona. However, it tends to cause overwriting, which harms the utility of the generated sentence, as indicated by the lowest \rouge score. The linear combination of scores introduces conflicts in selecting the best examples, leading to worse results. This is expected, as a high \privacyLeakage score only ensures strong privacy preservation but does not necessarily reflect utility. Therefore, the linear combination is not a feasible approach for scoring examples.

\paragraph{RQ 3: How is the quality of \privacyseg extraction?}
In our rewrite evaluation, we assume that the \privacyseg exactly matches the privacy specification for a more comprehensive evaluation of rewrite quality. However, in real-world scenarios, we first need to identify and align the \privacyseg before rewriting. To evaluate this component, we tested three possible methods. We considered using \cosine similarity and \armo for privacy token selection. Inspired by \cite{dou-etal-2024-reducing}, we also fine-tuned a \llamatwo model to detect privacy spans based on the persona. 
The results are shown in Table~\ref{tab::alignement_option}. For the first two approaches, we applied a token-level scoring method, setting a threshold of 0.2 for \cosine similarity and 0.15 for \armo, which were determined from the training set. Additionally, we evaluated the overlap coefficient~\cite{vijaymeena2016survey} for detected \privacyseg overlap using the following formula:$ \text{OC}(A, B) = \frac{|A \cap B|}{\min(|A|, |B|)} $ which measures the number of common tokens between two sets. The results show that fine-tuning \llamatwo achieves the best \rouge 89.38\% and \oc 88.88\%. With sufficient training data, the fine-tuned model performs well in span detection. On the other hand, metric-based methods require an appropriate cutoff threshold, making them less accurate compared to a fine-tuned model.

\paragraph{RQ4: Can our method effectively prevent reconstruction attacks for text rewriting?} A critical concern is whether an adversary, knowing the algorithm and accessing rewritten texts, can infer original private information. Using the optimal reconstruction framework from \cite{tong2025vulnerability}, we compute ASR bounds at the token level for differing tokens in $u$ and $y$, with token alignment for length discrepancies. The 3.07\% ASR in both context-free and contextual Bayesian attacks indicates robust protection, supporting the adaptability of \ourmethod to diverse privacy specifications without sacrificing coherence or relevance.

\paragraph{Cost latency analysis for performance.}
As a practical complement to our accuracy results, we provide an \emph{estimated} compute–cost and latency analysis using the same setup, together with a simple performance-per-unit-cost summary; see App.~\ref{app:cost}. 
This analysis is intended only as an order-of-magnitude estimate to contextualize deployment trade-offs.

\section{Related Work}
\paragraph{Privacy in Large Language Models}
Recent research highlights growing concerns over privacy risks in large-scale language models, where both explicit and implicit private information can be inferred from text generation \cite{brown2020language, dou-etal-2024-reducing}. Attack methods such as membership inference \cite{shokri2017membership} and reconstruction attacks \cite{lukas2023analyzing} reveal that models can memorize and leak sensitive details from training data. Prior studies have explored differential privacy mechanisms \cite{igamberdiev2023dp, bo2019er}, adversarial training \cite{barrett2019adversarial}, and explicit text anonymization \cite{akbik2019flair, lison2021anonymisation} to mitigate these risks given various context, but these methods often degrade text utility or limited in the certain form of privacy requirements. 
\paragraph{Privacy-Preserving Text Rewriting}
Privacy-aware text rewriting approaches typically rely on rule-based scrubbing \cite{akbik2019flair}, fine-tuned anonymization models \cite{dou-etal-2024-reducing}, or zero-shot prompting techniques \cite{utpala2023locally}. Rule-based approaches are precise but limited in flexibility, while fine-tuned models require extensive human supervision. Recent work has leveraged prompt engineering for privacy preservation without retraining models, demonstrating effectiveness in document rewriting \cite{meisenbacher2024dp}. \citet{staab2024large} considers LLM as an adversary to give feedback to the rewrite model to minimize the re-identification risk. Our study goes another direction to rewrite via tree-based rewrite with an explicit rewrite strategy.
\paragraph{Tree Search and Reward-Guided Generation} 
Tree search techniques have been increasingly explored in LLM-controlled text generation, allowing structured decision-making over multiple reasoning and generation paths. Tree of Thoughts (ToT) enables LLMs to explore multiple reasoning paths systematically, improving complex task-solving by iterating over various candidate solutions \cite{yao2023tree}. Similarly, Self-Play with Tree Search Refinement (SPaR) enhances model instruction by refining generated outputs through structured search and iterative decision-making \cite{cheng2024refinement}. Our method builds upon these principles by tree search and iterative refinement for privacy-preserving rewriting, ensuring progressive modifications that balance privacy, naturalness, and semantic preservation.
\section{Conclusion}
This paper introduces \ourmethod, a zero-shot tree-search based iterative privacy-aware rewriting method that adopts a MCTS-inspired search strategy. Through our extensive experiments, we show that our approach significantly outperforms baselines in terms of improved preservation of privacy, utility and naturalness. The MCTS-inspired search strategy is also superior to alternative methods. One possible direction is to further adapt and optimize \ourmethod as a versatile and generalized user privacy rewrite solution, particularly for on-device LLMs, to better accommodate evolving data release scenarios and granular user preferences.

\section{Limitations}
While \ourmethod effectively removes sensitive information and improves controllability, it has several limitations. First, the approach relies on model-driven rewriting, which may still retain implicit privacy cues or introduce inconsistencies due to the inherent variability of zero-shot prompting. Additionally, our method primarily focuses on general textual data, but privacy risks vary across formats such as emails, chat messages, and structured documents. Expanding the framework to context-aware privacy preservation could improve adaptability across different communication settings.

Second, due to budgetary constraints, our method's implementation was limited in scope, preventing large-scale human annotation for diverse rewriting strategies. While the dataset is sufficient to validate our findings, it may not generalize to all real-world privacy scenarios, particularly in commercial settings. Future work could explore generalized expansion via prompt tuning or more efficient algorithms to reduce computational costs. Moreover, our evaluation primarily relies on automatic metrics. Developing more refined privacy evaluation metrics that better align with human preferences presents a promising direction for future research.
\bibliography{main}

\begin{thebibliography}{35}
\providecommand{\natexlab}[1]{#1}

\bibitem[{Achiam et~al.(2023)Achiam, Adler, Agarwal, Ahmad, Akkaya, Aleman, Almeida, Altenschmidt, Altman, Anadkat et~al.}]{achiam2023gpt}
Josh Achiam, Steven Adler, Sandhini Agarwal, Lama Ahmad, Ilge Akkaya, Florencia~Leoni Aleman, Diogo Almeida, Janko Altenschmidt, Sam Altman, Shyamal Anadkat, et~al. 2023.
\newblock Gpt-4 technical report.
\newblock \emph{arXiv preprint arXiv:2303.08774}.

\bibitem[{Akbik et~al.(2019)Akbik, Bergmann, Blythe, Rasul, Schweter, and Vollgraf}]{akbik2019flair}
Alan Akbik, Tanja Bergmann, Duncan Blythe, Kashif Rasul, Stefan Schweter, and Roland Vollgraf. 2019.
\newblock Flair: An easy-to-use framework for state-of-the-art nlp.
\newblock In \emph{Proceedings of the 2019 conference of the North American chapter of the association for computational linguistics (demonstrations)}, pages 54--59.

\bibitem[{Barrett et~al.(2019)Barrett, Kementchedjhieva, Elazar, Elliott, and S{\o}gaard}]{barrett2019adversarial}
Maria Barrett, Yova Kementchedjhieva, Yanai Elazar, Desmond Elliott, and Anders S{\o}gaard. 2019.
\newblock Adversarial removal of demographic attributes revisited.
\newblock In \emph{Proceedings of the 2019 Conference on Empirical Methods in Natural Language Processing and the 9th International Joint Conference on Natural Language Processing (EMNLP-IJCNLP)}, pages 6331--6336.

\bibitem[{Bo et~al.(2019)Bo, Ding, Fung, and Iqbal}]{bo2019er}
Haohan Bo, Steven~HH Ding, Benjamin Fung, and Farkhund Iqbal. 2019.
\newblock Er-ae: differentially-private text generation for authorship anonymization.
\newblock \emph{arXiv preprint arXiv:1907.08736}.

\bibitem[{Brown(2020)}]{brown2020language}
Tom~B Brown. 2020.
\newblock Language models are few-shot learners.
\newblock \emph{arXiv preprint arXiv:2005.14165}.

\bibitem[{Chalkidis et~al.(2019)Chalkidis, Androutsopoulos, and Aletras}]{chalkidis2019neural}
Ilias Chalkidis, Ion Androutsopoulos, and Nikolaos Aletras. 2019.
\newblock Neural legal judgment prediction in english.
\newblock In \emph{Proceedings of the 57th Annual Meeting of the Association for Computational Linguistics}, pages 4317--4323.

\bibitem[{Chalkidis et~al.(2020)Chalkidis, Fergadiotis, Malakasiotis, Aletras, and Androutsopoulos}]{chalkidis-etal-2020-legal}
Ilias Chalkidis, Manos Fergadiotis, Prodromos Malakasiotis, Nikolaos Aletras, and Ion Androutsopoulos. 2020.
\newblock \href {https://doi.org/10.18653/v1/2020.findings-emnlp.261} {{LEGAL}-{BERT}: The muppets straight out of law school}.
\newblock In \emph{Findings of the Association for Computational Linguistics: EMNLP 2020}, pages 2898--2904, Online. Association for Computational Linguistics.

\bibitem[{Cheng et~al.(2024)Cheng, Liu, Wang, Gu, Lu, Zhang, Dong, Tang, Wang, and Huang}]{cheng2024refinement}
Jiale Cheng, Xiao Liu, Cunxiang Wang, Xiaotao Gu, Yida Lu, Dan Zhang, Yuxiao Dong, Jie Tang, Hongning Wang, and Minlie Huang. 2024.
\newblock \href {https://arxiv.org/abs/2412.11605} {Spar: Self-play with tree-search refinement to improve instruction-following in large language models}.
\newblock \emph{Preprint}, arXiv:2412.11605.

\bibitem[{Dainese et~al.(2024)Dainese, Merler, Alakuijala, and Marttinen}]{dainese2024generating}
Nicola Dainese, Matteo Merler, Minttu Alakuijala, and Pekka Marttinen. 2024.
\newblock Generating code world models with large language models guided by monte carlo tree search.
\newblock \emph{arXiv preprint arXiv:2405.15383}.

\bibitem[{Deuber et~al.(2023)Deuber, Keuchen, and Christin}]{deuber2023assessing}
Dominic Deuber, Michael Keuchen, and Nicolas Christin. 2023.
\newblock Assessing anonymity techniques employed in german court decisions: A $\{$De-Anonymization$\}$ experiment.
\newblock In \emph{32nd USENIX Security Symposium (USENIX Security 23)}, pages 5199--5216.

\bibitem[{Dou et~al.(2024)Dou, Krsek, Naous, Kabra, Das, Ritter, and Xu}]{dou-etal-2024-reducing}
Yao Dou, Isadora Krsek, Tarek Naous, Anubha Kabra, Sauvik Das, Alan Ritter, and Wei Xu. 2024.
\newblock \href {https://doi.org/10.18653/v1/2024.acl-long.741} {Reducing privacy risks in online self-disclosures with language models}.
\newblock In \emph{Proceedings of the 62nd Annual Meeting of the Association for Computational Linguistics (Volume 1: Long Papers)}, pages 13732--13754, Bangkok, Thailand. Association for Computational Linguistics.

\bibitem[{Du et~al.(2023)Du, Yue, Chow, Wang, Huang, and Sun}]{du2023dpforward}
Minxin Du, Xiang Yue, Sherman~SM Chow, Tianhao Wang, Chenyu Huang, and Huan Sun. 2023.
\newblock Dp-forward: Fine-tuning and inference on language models with differential privacy in forward pass.
\newblock In \emph{Proceedings of the 2023 ACM SIGSAC Conference on Computer and Communications Security}, pages 2665--2679.

\bibitem[{Huang et~al.(2024)Huang, MacLean, Kang, Wu, Qu, Xu, Li, Yuan, and Haffari}]{huang2024nap}
Shuo Huang, William MacLean, Xiaoxi Kang, Anqi Wu, Lizhen Qu, Qiongkai Xu, Zhuang Li, Xingliang Yuan, and Gholamreza Haffari. 2024.
\newblock Nap\^{} 2: A benchmark for naturalness and privacy-preserving text rewriting by learning from human.
\newblock \emph{arXiv preprint arXiv:2406.03749}.

\bibitem[{Igamberdiev and Habernal(2023)}]{igamberdiev2023dp}
Timour Igamberdiev and Ivan Habernal. 2023.
\newblock Dp-bart for privatized text rewriting under local differential privacy.
\newblock \emph{arXiv preprint arXiv:2302.07636}.

\bibitem[{Kocsis and Szepesv{\'a}ri(2006)}]{kocsis2006bandit}
Levente Kocsis and Csaba Szepesv{\'a}ri. 2006.
\newblock Bandit based monte-carlo planning.
\newblock In \emph{European conference on machine learning}, pages 282--293. Springer.

\bibitem[{Lison et~al.(2021)Lison, Pil{\'a}n, S{\'a}nchez, Batet, and {\O}vrelid}]{lison2021anonymisation}
Pierre Lison, Ildik{\'o} Pil{\'a}n, David S{\'a}nchez, Montserrat Batet, and Lilja {\O}vrelid. 2021.
\newblock Anonymisation models for text data: State of the art, challenges and future directions.
\newblock In \emph{Proceedings of the 59th Annual Meeting of the Association for Computational Linguistics and the 11th International Joint Conference on Natural Language Processing (Volume 1: Long Papers)}, pages 4188--4203.

\bibitem[{Lukas et~al.(2023)Lukas, Salem, Sim, Tople, Wutschitz, and Zanella-B{\'e}guelin}]{lukas2023analyzing}
Nils Lukas, Ahmed Salem, Robert Sim, Shruti Tople, Lukas Wutschitz, and Santiago Zanella-B{\'e}guelin. 2023.
\newblock Analyzing leakage of personally identifiable information in language models.
\newblock In \emph{2023 IEEE Symposium on Security and Privacy (SP)}, pages 346--363. IEEE.

\bibitem[{Meisenbacher et~al.(2024)Meisenbacher, Chevli, Vladika, and Matthes}]{meisenbacher2024dp}
Stephen Meisenbacher, Maulik Chevli, Juraj Vladika, and Florian Matthes. 2024.
\newblock Dp-mlm: Differentially private text rewriting using masked language models.
\newblock In \emph{Findings of the Association for Computational Linguistics ACL 2024}, pages 9314--9328.

\bibitem[{Mireshghallah et~al.(2023)Mireshghallah, Kim, Zhou, Tsvetkov, Sap, Shokri, and Choi}]{mireshghallah2023can}
Niloofar Mireshghallah, Hyunwoo Kim, Xuhui Zhou, Yulia Tsvetkov, Maarten Sap, Reza Shokri, and Yejin Choi. 2023.
\newblock Can llms keep a secret? testing privacy implications of language models via contextual integrity theory.
\newblock \emph{arXiv preprint arXiv:2310.17884}.

\bibitem[{{NVIDIA Corporation}(2022)}]{nvidia_6000}
{NVIDIA Corporation}. 2022.
\newblock {NVIDIA RTX-6000 GPU}.
\newblock \url{https://www.nvidia.com/en-au/design-visualization/rtx-6000/}.

\bibitem[{{NVIDIA Corporation}(2026)}]{nvidia_rtx5070ti}
{NVIDIA Corporation}. 2026.
\newblock {NVIDIA GeForce RTX 4070 Ti}.
\newblock \url{https://www.nvidia.com/en-au/geforce/graphics-cards/50-series/rtx-5070-family/}.

\bibitem[{Pan et~al.(2024)Pan, Lan, Li, and Qian}]{pan2024unsupervised}
Lei Pan, Yunshi Lan, Yang Li, and Weining Qian. 2024.
\newblock Unsupervised text style transfer via llms and attention masking with multi-way interactions.
\newblock \emph{arXiv preprint arXiv:2402.13647}.

\bibitem[{Raffel et~al.(2020)Raffel, Shazeer, Roberts, Lee, Narang, Matena, Zhou, Li, and Liu}]{raffel2020t5}
Colin Raffel, Noam Shazeer, Adam Roberts, Katherine Lee, Sharan Narang, Michael Matena, Yanqi Zhou, Wei Li, and Peter~J Liu. 2020.
\newblock Exploring the limits of transfer learning with a unified text-to-text transformer.
\newblock \emph{Journal of Machine Learning Research}, 21:1--67.

\bibitem[{Shokri et~al.(2017)Shokri, Stronati, Song, and Shmatikov}]{shokri2017membership}
Reza Shokri, Marco Stronati, Congzheng Song, and Vitaly Shmatikov. 2017.
\newblock Membership inference attacks against machine learning models.
\newblock In \emph{2017 IEEE symposium on security and privacy (SP)}, pages 3--18. IEEE.

\bibitem[{shunzh(2024)}]{mctsforllm}
shunzh. 2024.
\newblock Monte-carlo tree search for large language models (mcts-for-llm).
\newblock \url{https://github.com/shunzh/mcts-for-llm}.
\newblock Commit 0785eda83e452be318780003c5c1b9821debfbdc, accessed 2025-09-08.

\bibitem[{Staab et~al.(2023)Staab, Vero, Balunovi{\'c}, and Vechev}]{staab2023beyond}
Robin Staab, Mark Vero, Mislav Balunovi{\'c}, and Martin Vechev. 2023.
\newblock Beyond memorization: Violating privacy via inference with large language models.
\newblock \emph{arXiv preprint arXiv:2310.07298}.

\bibitem[{Staab et~al.(2024)Staab, Vero, Balunovi{\'c}, and Vechev}]{staab2024large}
Robin Staab, Mark Vero, Mislav Balunovi{\'c}, and Martin Vechev. 2024.
\newblock Large language models are advanced anonymizers.
\newblock \emph{arXiv preprint arXiv:2402.13846}.

\bibitem[{Tong et~al.(2025)Tong, Chen, Yuan, Liu, Zhang, Yu, and Zhang}]{tong2025vulnerability}
Meng Tong, Kejiang Chen, Xiaojian Yuan, Jiayang Liu, Weiming Zhang, Nenghai Yu, and Jie Zhang. 2025.
\newblock On the vulnerability of text sanitization.
\newblock In \emph{Proceedings of the 2025 Conference of the Nations of the Americas Chapter of the Association for Computational Linguistics: Human Language Technologies (Volume 1: Long Papers)}, pages 5150--5164.

\bibitem[{Utpala et~al.(2023)Utpala, Hooker, and Chen}]{utpala2023locally}
Saiteja Utpala, Sara Hooker, and Pin-Yu Chen. 2023.
\newblock Locally differentially private document generation using zero shot prompting.
\newblock In \emph{Findings of the Association for Computational Linguistics: EMNLP 2023}, pages 8442--8457.

\bibitem[{Vijaymeena and Kavitha(2016)}]{vijaymeena2016survey}
MK~Vijaymeena and K~Kavitha. 2016.
\newblock A survey on similarity measures in text mining.
\newblock \emph{Machine Learning and Applications: An International Journal}, 3(2):19--28.

\bibitem[{Wang et~al.(2024{\natexlab{a}})Wang, Xiong, Xie, Zhao, and Zhang}]{wang2024armo}
Haoxiang Wang, Wei Xiong, Tengyang Xie, Han Zhao, and Tong Zhang. 2024{\natexlab{a}}.
\newblock \href {https://arxiv.org/abs/2406.12845} {Interpretable preferences via multi-objective reward modeling and mixture-of-experts}.
\newblock \emph{Preprint}, arXiv:2406.12845.

\bibitem[{Wang et~al.(2024{\natexlab{b}})Wang, Xiong, Xie, Zhao, and Zhang}]{wang2024interpretable}
Haoxiang Wang, Wei Xiong, Tengyang Xie, Han Zhao, and Tong Zhang. 2024{\natexlab{b}}.
\newblock Interpretable preferences via multi-objective reward modeling and mixture-of-experts.
\newblock \emph{arXiv preprint arXiv:2406.12845}.

\bibitem[{Williams et~al.(2018)Williams, Nangia, and Bowman}]{mnli}
Adina Williams, Nikita Nangia, and Samuel Bowman. 2018.
\newblock \href {http://aclweb.org/anthology/N18-1101} {A broad-coverage challenge corpus for sentence understanding through inference}.
\newblock In \emph{Proceedings of the 2018 Conference of the North American Chapter of the Association for Computational Linguistics: Human Language Technologies, Volume 1 (Long Papers)}, pages 1112--1122. Association for Computational Linguistics.

\bibitem[{Yao et~al.(2023)Yao, Yu, Zhao, Shafran, Griffiths, Cao, and Narasimhan}]{yao2023tree}
Shunyu Yao, Dian Yu, Jeffrey Zhao, Izhak Shafran, Tom Griffiths, Yuan Cao, and Karthik Narasimhan. 2023.
\newblock \href {https://proceedings.neurips.cc/paper_files/paper/2023/file/271db9922b8d1f4dd7aaef84ed5ac703-Paper-Conference.pdf} {Tree of thoughts: Deliberate problem solving with large language models}.
\newblock In \emph{Advances in Neural Information Processing Systems}, volume~36, pages 11809--11822. Curran Associates, Inc.

\bibitem[{Zhang et~al.(2018)Zhang, Dinan, Urbanek, Szlam, Kiela, and Weston}]{zhang2018personaChat}
Saizheng Zhang, Emily Dinan, Jack Urbanek, Arthur Szlam, Douwe Kiela, and Jason Weston. 2018.
\newblock Personalizing dialogue agents: I have a dog, do you have pets too?
\newblock In \emph{Proceedings of the 56th Annual Meeting of the Association for Computational Linguistics (Volume 1: Long Papers)}, pages 2204--2213.

\end{thebibliography}
\newpage
\appendix
\section{appendix}
\label{sec:appendix}
\subsection{Baseline Method Implementation}\label{appendix::baseline}

\paragraph{\dpprompt.}\citet{utpala2023locally} utilizes zero-shot prompting and large language model to generate document paraphrasing to prevent author de-anonymization attack which comprise the privacy of text owner. In our tasks, as our backbone model is \llamato, we also use it as \dpprompt backbone. For $\epsilon$, we set it to 100 which is best empirical to balance all metric tested. 
\paragraph{\dpmlm} \citet{meisenbacher2024dp} considers BERT and Masked Language Prediction to gather improve the word utility for the generation distribution which bring improved utility in resulting generation. We maintain the same $\epsilon$ level to 100 to have comparative with other method.
\paragraph{\disclosure} \citet{dou-etal-2024-reducing} The self-disclosure abstraction detects the self disclosure span within given input text and perform rewrite using finetuned \llamato model. As their defined privacy is close to the one defined in \ourBenchmark, we directly adapts the rewrite model and test the generated result via our metric.
\paragraph{\gptf} The most powerful commercial language model\cite{achiam2023gpt}, we used the same prompt template from \cite{huang2024nap}  to generate strategy specific rewrite as one of the grouding baseline for our method.

\paragraph{\tfive-\ourBenchmark.} The SOTA privacy rewriting model based on \tfivesbase. It is fine-tuned on the dataset of \ourBenchmark directly with original utterance and human rewrite yielding the best privacy preserving score among all method 
\subsection{Upper Confidence Bound for Trees}\label{app:uct}
To guide the exploration of candidate rewrites during our \ourmethod, we adopt the Upper Confidence Bound for Trees (UCT) algorithm. UCT balances exploitation of high-reward candidates with exploration of less-visited options by selecting actions that maximize the following objective: 
\[
\text{UCT}(i) = \bar{X}_i + C \cdot \sqrt{\frac{\ln N}{n_i}},
\]
where $\bar{X}_i$ is the average reward of node $i$, $n_i$ is the number of times node $i$ has been visited, $N$ is the total number of visits to the parent node, and $C$ is a tunable exploration constant. In our setting, this mechanism allows us to prioritize rewrite trajectories that yield high reward scores while still exploring diverse rewriting paths. We set $C$ empirically as 6.36 based on validation performance to ensure sufficient exploration during tree expansion.
\subsection{Implementation Detail}
We conducted our experiments on a single A40 GPU\cite{nvidia_6000} with 46GB RAM, ensuring efficient execution of our tree-search-based privacy rewriting method. And we also  tested that our method can be run in GTX5070ti with 11GB RAM\cite{nvidia_rtx5070ti}.

The implementation of \ourmethod is adapted from the open-source repository Our MCTS decoder is adapted from the open-source \textit{mcts-for-llm}\footnote{The code is available at \url{https://github.com/shunzh/mcts-for-llm}.} implementation \citep{mctsforllm}.
, released under MIT. We adopted the frame for MCTS decoding and convert it to sentence level generation and implement our algorithm based on the framework. 
To control the computational overhead, we set the tree search computation budget to 5, allowing iterative refinement while maintaining feasible inference times. For model generation, we adopted a top-$p$ probability of 5, ensuring diverse sampling while maintaining high-quality outputs. The maximum generation length was constrained to 128 tokens to prevent excessive expansion and maintain sentence coherence.

We set threshold for our reward model empirically to 0.10 to filter the rewrite quality. This threshold is obtained via training set to obtain best performance. Inference on our test set (280 instances) required approximately 2 hours, while processing the \echr dataset took significantly longer, requiring 23 hours due to the complexity of legal text and entity alignment. In our experiment of \echr dataset, the consumption is significant for some reasons. As the ECHR dataset is constructed by cases which contains 109 sentences per case with 20.33 words per sentence. By contract, each example in NAP is just one sentence with 140 examples with 14.25 tokens per examples which is more natural for human conversations. Thus the long running time for ECHR is acceptable under this circumstances in our experiment. It is noted that the generation with more inference steps raises significant computation overhead and latency. It also raises the further direction of optimizing such generation approach to minimize the inference cost.

\subsection{Limitation for LLMs} In our work, the spotted weakness for privacy rewrite of LLM in one-off rewrite particular happens in the scenario where the sentence requires rewrite based on persona, for example if user ask to rewrite sentence "I am an Ohio Mon with two amazing sons,  not married though" based on persona "I am a single mom of two boys". The SOTA LLM often fails to rewrite all possible private information mentioned in the persona as their alignment have different focus in multiple target. For GPT-4 model, the rewrite goes to "I live in Ohio and have a wonderful family, but I'm not married" and for open source model which is less competitive the result goes worse while handling complex privacy rewrite. Thus, we argue to decompose the private information for sentence to obtain better rewrite. 

\subsection{Additional Utility Metrics}
\label{app:utility-metrics}

\textbf{Scope.} These measurements provide supplementary evidence about text variety and semantic alignment. 
They are \emph{complements} to our primary utility/accuracy metrics and should be interpreted accordingly.

\textbf{Definitions.} 
(1) \emph{Diversity (Distinct-2)} = \#unique bigrams / \#total bigrams (computed per sample and averaged); 
(2) \emph{MAUVE} measures distributional similarity between the model rewrites and references (higher is better); 
(3) \emph{SimCSE} is cosine similarity between sentence embeddings of rewrite and reference (higher is better).

\textbf{Setup.} All methods are evaluated on the same test set and decoding configuration as the main experiments; preprocessing and tokenization follow the respective reference implementations of each metric.

\textbf{Notes and caveats.} 
Diversity and MAUVE can be sensitive to generation length/temperature; SimCSE depends on the encoder backbone. 
These indicators are provided to triangulate quality, not to replace task-specific utility or privacy outcomes.
\begin{table*}[t]
\centering

\small
\setlength{\tabcolsep}{8pt}
\begin{tabular}{lccc}
\toprule
Method & Diversity (Distinct-2) & MAUVE & SimCSE \\
\midrule
\dpmlm & 2.5882 & 0.2391 & 0.4888 \\
\dpprompt & 2.9810 & 0.0044 & 0.1719 \\
\flair & 1.1999 & 0.0170 & 0.6572 \\
\disclosure & 2.0976 & 0.0725 & 0.2987 \\
\gptf (one-pass) & 1.3521 & \textbf{0.5878} & \textbf{0.5691} \\
\tfive-\ourBenchmark-\gptf & 0.5932 & 0.1794 & 0.2696 \\
\ourmethod-\llamato & \textbf{2.1293} & 0.0235 & 0.5225 \\
\bottomrule
\end{tabular}
\caption{Additional utility metrics (higher is better). 
Diversity is Distinct-2; SimCSE is cosine similarity to the reference. 
MAUVE compares the distributions of references and rewrites.}
\label{tab:utility_extra}
\end{table*}

\subsection{Estimated Cost–Latency–Performance Analysis}
\label{app:cost}

\textbf{Goal.} We estimate how our method trades off privacy performance against compute cost and latency under the same evaluation setup. 
This is \emph{not} a billing statement; all numbers are approximate and depend on hardware, rates, batching, and provider pricing.

\paragraph{Definitions.}
Let $P_{\text{ours}}, P_{\text{base}}$ be privacy scores on the test set (higher is better), and 
$C_{\text{ours}}, C_{\text{base}}$ be the estimated compute cost per 100 examples under identical conditions.
We report the \emph{incremental performance gain}
$\Delta P = P_{\text{ours}}-P_{\text{base}}$
and the \emph{performance-per-unit-cost}
\[
\frac{\Delta P}{\Delta C}
\;=\;
\frac{P_{\text{ours}}-P_{\text{base}}}{\,C_{\text{base}}-C_{\text{ours}}\,} \]
This normalizes incremental benefit by incremental (net) cost on the same task, which is appropriate when contrasting alternative implementations of the \emph{same} functionality.

\paragraph{Concrete figures.}
We use \textsc{LLaMA~3.1–8B} on an \textsc{A40} GPU.
Processing 100 examples takes \textbf{42.5 minutes} ($\approx$\,\textbf{51 s}/sentence). 
At a \emph{nominal} rental rate of \$\textbf{0.47}/hour, this yields 
$C_{\text{ours}}=\$\,\mathbf{0.332}$ (=$0.708\text{ h}\times\$0.47/\text{h}$).
For a one-pass GPT-4 baseline via API we use $C_{\text{base}}=\$\,\mathbf{0.42}$ per 100 examples.
Measured privacy scores are $P_{\text{ours}}=\mathbf{93.02}$ and $P_{\text{base}}=\mathbf{82.24}$, so $\Delta P=\mathbf{10.78}$.
Therefore,
\[
\frac{\Delta P}{\Delta C}
=
\frac{10.78}{0.42-0.332}
=
\frac{10.78}{0.088}
=\mathbf{122.5}\;
\]

\paragraph{Scope and caveats.}
All values are estimates; real costs vary with GPU market rates, region, utilization, batch size, tokenization, prompt length, caching, parallelization, and API pricing tiers. 
We exclude engineering time, storage, networking, and overheads. 
Latency reflects a single-GPU desktop/offline configuration; cloud inference, multi-GPU parallelism, or quantization can materially change results.
We encourage readers to recompute with their own rates using the formulas above.

\begin{table*}[t]
\centering
\setlength{\tabcolsep}{6pt}
\begin{tabular}{lcccccc}
\toprule
Method & $P$ & $\Delta P$ & Time (100 ex.) & Cost/100 ex. & $\Delta C$ & $\Delta P/\Delta C$ \\
\midrule
Ours (\textsc{LLaMA~3.1–8B}, A40) & 93.02 & \multirow{2}{*}{10.78} & 42.5 min & \$\,0.332 & \multirow{2}{*}{0.088} & \multirow{2}{*}{122.5} \\
Baseline (GPT-4, one-pass)        & 82.24 &                       & --       & \$\,0.42  &                         & \\
\bottomrule
\end{tabular}
\caption{Estimated cost–latency–performance summary for 100 examples.
$\Delta P/\Delta C$ is in \emph{points per dollar}. 
Figures are approximate and for contextual comparison only.}
\label{tab:cost_summary}
\end{table*}

\paragraph{Reproduction details.}
For open-weight inference we compute $C_{\text{ours}}=r \times t$ with $r=\$0.47$/h and $t=0.708$\,h for 100 examples; 
for API we use a per-100-example estimate of \$0.42 aligned to our prompt/response lengths.
Please substitute your own $r$ and token pricing to recompute locally.

\subsection{Detailed explanation of the example}

We consider the rewrite example with input sentence "I am an Ohio Mon with two amazing sons,  not married though" and persona information "I am a single mom of two boys." As shown in Figure.~\ref{rewrit_example}. From the datasets. There will be four parts of token considered as private segments based on persona. In this case we conduct token-level mapping ['mom' 'two' 'sons' 'Not married'] For each private segments, we conduct our sentence level privacy rewrite in tree search with either obscuring or deleting strategies. The node randomly takes action of \deleting at the beginning resulting in the output "I am an Ohio with two amazing sons, not married though". The reward function evaluates the rewrite and decides to take the next action of \deleting in original input sentence. The subsequent output "I am an Ohio resident with two amazing sons, not married" are chosen via empirically set threshold for 0.10. It is accepted as partial rewritten sentence for next privacy segment. After that the next private segment "two" will be deleted. After all rewrites, the final output goes to  "I am an Ohio resident with amazing children". 

\subsection{\armo}\label{app:armo}
We adopt \armo \cite{wang2024interpretable} as the reward model guiding our tree-search-based rewriting process. \armo is a state-of-the-art preference modeling framework that learns multi-objective reward functions from human feedback through a mixture-of-experts design. It produces interpretable reward signals aligned with human judgment across multiple dimensions, such as coherence, relevance, and instruction-following quality. We select \armo due to its strong empirical performance in evaluating text generation quality and its modular design that allows fine-grained control over different objectives during rewriting. They incorporate the Mixture-of-Experts (MoE) gating mechanism that selects objective-specific weights dynamically depending on the context (prompt), allowing adaptive and steerable reward aggregation. In our setting, the reward score generated by \armo is used to evaluate each candidate rewrite during the tree search process. Emperical result in comparsion shows it works

\subsection{LLM as a judge evaluation of naturalness}
\label{app:naturalness}
We employ GPT-4o as LLM judge to score the generated output. As shown in Table.~\ref{tab:baseline}, each example using template in Fig.~\ref{fig:llm-naturalness-prompt}. The score ranges from 1 to 5. 1 denotes very unnatural of result and 5 denotes the high fluent and native sounding of a rewritten sentences. We set the human rewrite as naturalness to set comparison baseline for all methods. 
\begin{figure*}[t]
\centering
\begin{tcolorbox}[
    colframe=promptborder,
    coltitle=white,
    colbacktitle=prompttitle,
    title=\textbf{LLM Naturalness Judgment Prompt},
    fonttitle=\bfseries,
    width=0.95\textwidth,
    boxrule=1pt,
    arc=3pt,
    sharp corners=southwest,
]
You are an expert linguist. Your task is to assess the naturalness of a given sentence — how fluent, human-like, and typical it sounds in everyday language use.

\vspace{0.5em}
\textbf{Rate the sentence on a scale from 1 to 5:}
\begin{itemize}
    \item \textbf{1} = very unnatural (awkward, grammatically incorrect, or robotic)
    \item \textbf{2} = mostly unnatural
    \item \textbf{3} = somewhat natural (acceptable but slightly awkward)
    \item \textbf{4} = mostly natural (minor issues)
    \item \textbf{5} =  very natural (fluent and native-sounding)
\end{itemize}

\vspace{0.5em}
\textbf{Sentence:} \verb|"Sentence to Assess"|

\vspace{0.5em}
Only provide the score and a brief explanation in the following JSON format:

\begin{verbatim}
{"score": X, "explanation": "..."}
\end{verbatim}
\end{tcolorbox}
\caption{Prompt used for LLM-based naturalness judgment.}
\label{fig:llm-naturalness-prompt}
\end{figure*}

\subsection{More examples with rewrite}
Table~\ref{tab:privacy_examples} displays 8 privacy rewriting examples from two domains. Examples 1–4 are from the \ourBenchmark, where each input sentence is accompanied by a user-provided persona sentence that conveys private specification. The rewriting task involves aligning the input and persona to identify segments for rewriting. These privacy spans are highlighted in bold within the sentence. The rewrites aim to generalize or obfuscate sensitive information while maintaining fluency and semantic coherence. In Example 3, we additionally show a human-authored rewrite for comparison. As per request that we assume user require to fully remove information from the sentence. In human judgment it requires fully remove of later part. In this case even information of disability is hided by model. It actually failed to give the desired rewrite like human rewrite.

Examples 5–8 are randomly drawn from the ECHR dataset, comprising legal case descriptions with sensitive attributes such as nationality, date of birth, location, and organization names. These were automatically scrubbed using the \flair, with the masked spans treated as private segments for rewriting. The rewritten outputs illustrate the model's ability to reconstruct fluent, semantically accurate paraphrases while replacing private entities with abstract or generalized descriptions. For instance, in Example 6, organizational and location-specific terms are replaced with more generic descriptors of institutions and international settings.

\begin{table*}[t]
\centering
\small
\begin{tabular}{p{0.95\textwidth}}
\toprule

\textit{\ourBenchmark Example 1} \\
\textbf{Sentence:} \textbf{Scotch} is my drink . Helps keep the evil dogs away , they scare me . \\
\textbf{Persona:} I like to drink \textbf{scotch} to relax. \\
\textbf{Rewrite:} My amber liquid of choice. Helps keep the menacing beasts at bay, they frighten me. \\

\midrule

\textit{\ourBenchmark Example 2} \\
\textbf{Sentence:} Hello . I live in an \textbf{apartment} . It is a \textbf{low income} residence . \\
\textbf{Persona:} I live in \textbf{low income} \textbf{apartments}.  \\
\textbf{Rewrite:} Hello. I reside in our community. \\

\midrule

\textit{\ourBenchmark Example 3} \\
\textbf{Sentence:} That is great . I am \textbf{unable} to work due to my \textbf{disability}. \\
\textbf{Persona:} I am unable to \textbf{work} at a \textbf{job} because of a \textbf{disability}.         \\
\textbf{Rewrite:} “That is great. I am currently experiencing certain challenges in my ability to work due to my personal circumstances.” \\
\textbf{Human Rewrite:} \textcolor{red}{That is great!} \\

\midrule

\textit{\ourBenchmark Example 4} \\
\textbf{Sentence:} I \textbf{did not graduate} from \textbf{high school}, so writing novels seems like a great achievement \! \\
\textbf{Persona:} I \textbf{never finished high school}.  \\
\textbf{Rewrite:} I achieved great things despite completing my academic journey, so crafting stories feels like a remarkable accomplishment. \\
\midrule

\textit{\echr Example 5} \\
\textbf{Sentence:} The applicant is a \textbf{British national}, born in \textbf{1945} and living in \textbf{Rome}. \\
\textbf{Masked:} The applicant is a <MASK> national, born in <MASK> and living in <MASK>. \\
\textbf{Rewrite:} The applicant is a native of a major European capital city, born during the post-war era and residing in the Eternal City. \\

\midrule

\textit{\echr Example 6} \\
\textbf{Sentence:} In \textbf{1982} the \textbf{first} applicant, a civil engineer by profession and employed by the \textbf{French} company S., was placed at the disposal of \textbf{the European Space Agency} to perform services at the \textbf{European Space Operations Centre} in \textbf{Darmstadt}. \\
\textbf{Masked:}  In <MASK> the <MASK> applicant, a civil engineer by profession and employed by the <MASK> company S., was placed at the disposal of <MASK> to perform services at <MASK> in <MASK>.  \\
\textbf{Rewrite:} "A trailblazing innovation specialist, with a background in urban development, was seconded to a prestigious intercontinental space research institution to provide expertise in a cosmopolitan hub. \\

\midrule

\textit{\echr Example 7} \\
\textbf{Sentence:} After repeated motions for bias filed by the applicant's husband, the \textbf{Buda Central District Court} was \textbf{first} appointed to hear the case. \\
\textbf{Masked:} After repeated motions for bias filed by the applicant's husband, <MASK> was <MASK> appointed to hear the case. \\
\textbf{Rewrite:} "After repeated motions for bias filed by the applicant's husband, a court was appointed to hear the case." \\

\midrule

\textit{\echr Example 8} \\
\textbf{Sentence:} By \textbf{two} separate judgments of \textbf{4 June 2007}, \textbf{one} in favour of Mr \textbf{Pavlenko} and another in favour of the remaining applicants, the \textbf{Mineralniye Vody Town Court} of the \textbf{Stavropol Region} upheld their actions in part. \\
\textbf{Masked:} By <MASK> separate judgments of <MASK>, <MASK> in favour of Mr <MASK> and another in favour of the remaining applicants, <MASK> of <MASK> upheld their actions in part. \\
\textbf{Rewrite:} Copies were sent to nearly two dozen other individuals in the government and the federal authorities and to the diplomatic missions in the coastal metropolis, the radiant capital, and the central city. \\

\bottomrule
\end{tabular}
\caption{8 randomly sampled examples generated with \ourmethod-\llamato. 4 examples from \ourBenchmark and 4 examples from \echr. The privacy segments are marked \textbf{bold}}
\label{tab:privacy_examples}
\end{table*}

\end{document}